\documentclass{article}

\usepackage{microtype}
\usepackage{graphicx}
\usepackage{subfigure}
\usepackage{booktabs} 

\usepackage{hyperref}

\usepackage[accepted]{icml2024}

\usepackage{amsmath}
\usepackage{amssymb}
\usepackage{mathtools}
\usepackage{amsthm}
\usepackage{amsfonts}

\usepackage[capitalize,noabbrev]{cleveref}

\theoremstyle{plain}

\theoremstyle{definition}

\theoremstyle{remark}

\usepackage[textsize=tiny]{todonotes}
\usepackage[utf8]{inputenc} 
\usepackage[T1]{fontenc}    
\usepackage{url}            
\usepackage{nicefrac}       
\usepackage{xcolor}         
\usepackage{soul}
\usepackage{xspace}
\usepackage{algorithm}
\renewcommand{\algorithmicrequire}{\textbf{Input:}}
\renewcommand{\algorithmicensure}{\textbf{Output:}}
\usepackage{array}
\usepackage{eqparbox}
\renewcommand\algorithmiccomment[1]{{\color{gray} #1}}

\usepackage{etoolbox}  
\makeatletter
\patchcmd{\algorithmic}{\addtolength{\ALC@tlm}{\leftmargin} }{\addtolength{\ALC@tlm}{\leftmargin}}{}{}
\makeatother

\newcommand{\cost}{\ell}
\newcommand{\seed}{\xi}
\newcommand{\param}{\phi}

\newcommand{\saaparam}{{\tilde{\phi}}}
\newcommand{\paramset}{\Phi}
\newcommand{\sample}{z}
\newcommand{\transform}{T_\param}

\newcommand{\vardist}{q}

\newcommand{\threshold}{\alpha}
\newcommand{\oursfull}{variational inference with sequential sample-average approximations }

\newcommand{\OursFull}{Variational Inference with Sequential Sample-Average Approximations}
\newcommand{\ours}{VISA\xspace}
\newcommand{\baseline}{IWFVI\xspace}
\newcommand{\Exp}{\mathop{\mathbb{E}}}

\begin{document}

\twocolumn[
\icmltitle{\ours: \OursFull}

\icmlsetsymbol{equal}{*}



\begin{icmlauthorlist}
\icmlauthor{Heiko Zimmermann}{uva}
\icmlauthor{Christian A. Naesseth}{uva}
\icmlauthor{Jan-Willem van de Meent}{uva}
\end{icmlauthorlist}

\icmlaffiliation{uva}{
Amsterdam Machine Learning Lab, 
University of Amsterdam, 
Amsterdam, The Netherlands
}
\icmlcorrespondingauthor{Heiko Zimmermann}{h.zimmermann@uva.nl}

\icmlkeywords{Variational Inference, Importance Sampling}

\vskip 0.3in
]



\printAffiliationsAndNotice{}  

\begin{abstract}
We present \oursfull (\ours), a method for approximate inference in computationally intensive models, such as those based on numerical simulations. \ours extends importance-weighted forward-KL variational inference by employing a sequence of sample-average approximations, which are considered valid inside a trust region. This makes it possible to reuse model evaluations across multiple gradient steps, thereby reducing computational cost. We perform experiments on high-dimensional Gaussians, Lotka-Volterra dynamics, and a Pickover attractor, which demonstrate that \ours can achieve comparable approximation accuracy to standard importance-weighted forward-KL variational inference with computational savings of a factor two or more for conservatively chosen learning rates.
\end{abstract}

\section{Introduction}

The development of general-purpose methods for Bayesian inference, such as those provided by probabilistic programming systems \cite{vandemeentIntroductionProbabilisticProgramming2021}, has made it possible to apply Bayesian analysis to simulation-based models in the natural and physical sciences. 
The reason that these methods are so broadly applicable is that they generate samples by way of repeated evaluation of the model. 
However, this generality often comes at significant computational cost. Inference may require thousands of model evaluations, where each evaluation may itself involve computationally intensive operations such as numerical integration of a differential equation, or a call to a numerical solver. 

Gradient-based methods have become the workhorse for inference in simulation-based models. When a model defines a fully differentiable density, methods based on Hamiltonian Monte Carlo \cite{hoffmanNoUTurnSamplerAdaptively2014a} and reparameterized variational inference \cite{kingmaAutoEncodingVariationalBayes2013, rezendeStochasticBackpropagationApproximate2014} are often considered the gold standard for generating high quality samples from the posterior distribution. 
However, it is not always practical or possible to use a differentiable model. The implementation of the simulator may not support differentiation, or the model itself may not be differentiable, for example because it employs discrete random variables or stochastic control flow. In such cases, inference falls back on methods based on score-function estimators \cite{glynnLikelihoodRatioGradient1990, wingate2013automated, ranganathBlackBoxVariational2014}  
or importance-weighted forward-KL variational inference (\baseline), which derives from reweighted wake-sleep methods \cite{bornscheinReweightedWakeSleep2015, leRevisitingReweightedWakeSleep2018}. These methods are less computationally efficient, but remain the most viable option in a substantial number of use cases. 

In this paper, we present \ours, a method that has the potential to substantially improve the computational efficiency of variational inference for models that are non-differentiable and computationally intensive. The key idea is that evaluation of the variational approximation will typically be cheap relative to that of the model. This means that we can save computation by reusing model evaluations across multiple updates of the variational posterior. 
To this end, we adapt \baseline to employ a series of Sample-Average Approximations (SAA), which use a fixed set of samples that defines a deterministic surrogate to the objective, rather than generating a fresh set of samples at each gradient step. 


SAA methods were recently studied in the context of reparameterized black-box variational inference \cite{giordanoBlackBoxVariational2023, burroniSampleAverageApproximation2023} which optimizes the reverse KL-divergence. These methods fix samples from a parameter-free distribution, which are transformed to samples from the approximate posterior using a differentiable map, whose parameters are optimized to maximize the variational bound. \ours differs from these methods in that it optimizes a forward-KL divergence and does not require a differentiable model. Concretely, \ours fixes samples from a parameterized variational distribution, rather than samples from a parameter-free distribution. Since the variational distribution will change during optimization, we construct a new SAA whenever the optimization leaves a trust region, which we define in terms of the effective sample size. \ours is a drop-in replacement for \baseline, in which samples are re-used as much as possible, thereby saving computation.

We evaluate \ours in the context of three experiments. We first consider high-dimensional Gaussians, where the approximation error can be computed exactly. We then consider inference in a Lotka-Volterra system and a Pickover Attractor, where numerical integration is performed as part of the forward simulation. Our results show that \ours with a conservative (i.e.~smaller than needed) step size can converge in a smaller number of model evaluations than IWFVI with a more carefully tuned step size. These results come with the caveat that \ours is more susceptible to bias than IWFVI, especially when used with a low effective sample size threshold.
In our experiments, savings of a factor two or more are realizable with conservatively chosen learning rates, while \ours performs on par with \baseline for more carefully tuned step sizes.

\section{Background}
We first briefly review VI with SGD and SAAs, before we introduce \ours in Section \ref{sec:methods}. Readers familiar with these topics can safely skip ahead.

\subsection{Variational Inference}

Variational Inference (VI) methods approximate an intractable target density with a tractable variational distribution by solving an optimization problem. The objective is typically to minimize a divergence measure $\mathcal{D}$ between the variational approximation $\vardist_\param$ with parameters $\param \in \paramset$ and the target density $\pi$,
\begin{align}
    \label{eq:vi_opt_problem}
    \min_{\param \in \paramset}
    \left\{
        L(\param)
        :=
        \mathcal{D}(\vardist_\param , \pi) + const.
    \right\}.
\end{align}

In this exposition, we assume that the target density is the posterior of a probabilistic model $\pi(z) = p(z \mid y)$ for which we are able to point-wise evaluate the joint density $p(y,z)$.

The two most common approaches to VI are to minimize the reverse or forward KL divergence, for which objectives can be defined in terms of a lower and upper bound on the log marginal $\log p(y)$,
\begin{align*}
    L_\textsc{r}(\param)
    &=
    -
    \Exp_{q_\param}\left[
        \log \frac{
            p(y, \sample)
        }{
            q_\param(\sample)
        }
    \right]
    \!=
    \text{KL}(q_\phi \,||\, p(\cdot | y)) - \log p(y) 
    ,
    \\
    L_\textsc{f}(\param)
    &=
    \!\!\!\!
    \Exp_{p(\cdot \mid y)}\!\!
    \left[
        \log \frac{p(y, \sample)}{\vardist_\param(\sample)}
    \right]
    \!=
    \text{KL}(p(\cdot | y) \,||\, q_\phi) + \log p(y)
    .
\end{align*}

We will briefly discuss standard reparameterized VI, which maximizes the lower bound $-L_\textsc{r}$, and \baseline, which minimizes the upper bound $L_\textsc{f}$. 


\paragraph{Reparameterized VI.} When maximizing a lower bound with stochastic gradient descent, we can either employ score-function estimators \cite{ranganathBlackBoxVariational2014, paisley2012variational}, which tend to exhibit a high degree of variance, or make use of the reparameterization trick, which is commonly employed in variational autoencoders \cite{kingmaAutoEncodingVariationalBayes2013, rezendeStochasticBackpropagationApproximate2014}. 

Reparameterization is a method for generating $z \sim q_\phi$ by way of a parametric pushforward of random variables $\xi$ that are distributed according to a parameter-free density $q_\xi$,
\begin{align*}
    z &= T_\phi(\xi), & \xi \sim q_\xi.
\end{align*}
In VI, reparameterized samples can be used to compute an unbiased estimate of the gradient,
\begin{align*}
    -\frac{d}{d \param}
    L_\textsc{r}(\param)
    &=
    \Exp_{q_\seed}\left[
        \frac{d}{d \param}
        \log \frac{
            p(y, \transform(\seed))
        }{
            q_\param(\transform(\seed))
        }
    \right]
    \\
    &\approx
    \frac{1}{N}
    \sum_{i=1}^N
        \frac{d}{d \param}
        \log \frac{
            p(y, \transform(\seed^{(i)}))
        }{
            q_\param(\transform(\seed^{(i)}))
        }
    ,&& 
    \seed^{(i)}\sim q_\seed
    .
\end{align*}
Reparameterized VI requires a model $p(y, z)$ that is differentiable with respect to $z$ in order to compute the pathwise derivative \cite{rezendeStochasticBackpropagationApproximate2014}, 
\begin{align*}
    \frac{d}{d\param} \log p(y, T_\phi(\xi))
    =
    \left. \frac{\partial}{\partial z} \log p(y, z) \right\vert_{z=T_\phi(\xi)}
    \frac{\partial T_\phi(\xi)}{\partial \phi}
    .
\end{align*}
This means that the model must support (automatic) differentiation and must not make use of discrete random variables or stochastic control flow, such as if statements that branch on random values \cite{vandemeentIntroductionProbabilisticProgramming2021}.


\paragraph{Importance-Weighted Forward-KL VI.}

When minimizing the forward KL divergence, we approximate the gradient 
\begin{align*}
    -
    \frac{d}{d\param}
    L_\textsc{f}(\param)
    =
    \Exp_{p(\cdot \mid y)}\!\!
    \left[
        \frac{d}{d\param}
        \log \vardist_\param(\sample)
    \right]
    .
\end{align*}
This does not require differentiability of the model $p(y,z)$ with respect to $z$, but does  require approximate inference to generate samples from the posterior $p(z \mid y)$.

In IWFVI, the variational distribution $q_\param$ is used as a proposal for a self-normalized importance sampler. This approach makes use of the fact that we can express the gradient as an expectation with respect to $q_\param$,
\begin{align*}
    -
    \frac{d}{d\param}
    L_\textsc{f}(\param)
    =
    \Exp_{q_\param}\!\!
    \left[
        w_\param(\sample)
        \frac{d}{d\param}
        \log \vardist_\param(\sample)
    \right]
    ,
    ~~
    w_\phi(z) = \frac{p(z \,|\, y)}{q_\phi(z)}.
\end{align*}
Here the ratio $w_\phi(z)$, known as the importance weight, needs to be approximated, since we cannot compute $p(z \mid y)$. To do so, we define the self-normalized weights
\begin{align*}
    \hat{w}_\phi^{(i)} &= \frac{\bar{w}_\phi^{(i)}}{\sum_{j=1}^N \bar{w}_\phi^{(j)}},
    &
    \bar{w}_\phi^{(i)} = \frac{p(y, z^{(i)})}{q_\phi(z^{(i)})},
\end{align*}
leading to the gradient estimate
\begin{align}
    \label{eq:iw_grad}
    \frac{d}{d\param}
    L_\textsc{f}(\param)
    \simeq
    \sum_{i=1}^N
    \hat w^{(i)}
    \frac{d}{d\param}
    \log q_\param(\sample^{(i)})
    ,&&
    z^{(i)} \sim q_\param.
\end{align}
The resulting estimate is biased but consistent, meaning that it converges to the true gradient as $N\to \infty$ almost surely.

\subsection{VI with Sample-Average Approximations}

In the stochastic optimization literature, sample-average approximations are used to approximate an expected loss with a surrogate loss in the form of a Monte Carlo estimate (see \citet{kimGuideSampleAverage2015} for a review). Importantly, the samples that the SAA is based on remain \emph{fixed} throughout the optimization process. This means that the surrogate objective can be treated like any other deterministic function,
which can be optimized using second-order methods and other standard optimization tools.

Concretely, a sample-average approximation applies to an optimization problem of the form 
\begin{align}
 \label{eq:opt_problem}
 \min_{\param \in \paramset}
     \left\{
     L(\param) 
     := 
     \Exp_{\rho}
     \left[
         \cost(\sample, \param)
     \right]
 \right\}
 ,
\end{align}
in which the density $\rho(z)$ does not depend on the parameters $\phi$. This means that we can compute a surrogate loss $\hat{L}(\phi)$ that is an unbiased estimate of the original loss $L(\phi)$ by averaging over samples from $\rho$,
\begin{align*}
    \hat{L}(\phi) &= \frac{1}{N} \sum_{i=1}^N l(z^{(i)}, \phi), 
    &
    z^{(i)} &\sim \rho.
\end{align*}
Under mild conditions on $\cost$ and $\rho$, as the number of samples $N \to \infty$, the minimizer $\hat\param = \arg\min_\param \hat{L}(\param)$ and the minimal value $\hat{L}(\hat\param)$ converge almost surely to the minima $\param^* = \arg\min{L}_\param(\param)$ and minimal value $L(\param^*)$ of the original problem. 

In the context of reparameterized VI, a sample-average approximation can be constructed by fixing a set of samples $\{\xi^{(i)} \sim q_\xi\}_{i=1}^N$ from a parameter-free distribution,
\begin{align*}
    \hat{L}_\textsc{r}(\phi) &= \frac{1}{N} \sum_{i=1}^N \log \frac{p\big(y, T_\phi(\xi^{(i)})\big)}{q_\phi\big(T_\phi(\xi^{(i)})\big)}
    .
\end{align*}
In an SAA-based approach to reparameterized VI \cite{giordanoBlackBoxVariational2023, burroniSampleAverageApproximation2023}, optimization of the parameters $\phi$ will move the transformed samples $z^{(i)}=T_\phi(\xi^{(i)})$ to match the posterior density, whilst keeping the noise realizations $\seed^{(i)}$ fixed. Empirical evaluations show that combining the SAA approximation with an off-the-shelf second-order optimizer can result in substantial computational gains as well as more reliable convergence to the optimum.

\section{SAA for Forward-KL Variational Inference}
\label{sec:methods}


The primary motivation behind existing SAA-based methods for reparameterized VI \citep{giordanoBlackBoxVariational2023, burroniSampleAverageApproximation2023} is that fixing the noise realizations defines a completely deterministic surrogate objective $\hat{L}_\textsc{r}(\phi)$, which can then be used with any number of existing optimizers. The main requirement from an implementation point of view is that the model density $p(y,z)$ is differentiable with respect to $z$. In this setting it is also necessary to evaluate the model for every update, since any change to $\phi$ also changes the values of the transformed samples $z^{(i)} = T_\phi(\xi^{(i)})$.

In developing \ours, both our motivation and implementation requirements are somewhat different. Our primary interest is in minimizing the total number of model evaluations at convergence. We also wish to develop a method that is applicable when the model density $p(y,z)$ is not differentiable, either because the implementation simply does not support (automatic) derivatives, or because the model incorporates discrete variables or stochastic control flow, which introduce discontinuities in the density $p(y,z)$.

To this end, we propose a method that optimizes a forward KL with an importance weighted objective 
that incorporates ideas from SAA-based approaches. In a setting where we already have access to samples from the posterior, we could trivially define a SAA for the upper bound $L_\textsc{f}(\phi)$, 
\begin{align*}
    \hat{L}_\textsc{f}(\phi) &= \frac{1}{N} \sum_{i=1}^N \log \frac{p(y, z^{(i)})}{q_\phi(z^{(i)})},
    &
    z^{(i)} &\sim p(\cdot \mid y).
\end{align*}
In practice, this naive approach is unlikely to be useful in a setting where evaluation of $p(y, z)$ is computationally expensive, since we would still need to carry out approximate inference to generate a set of samples from the posterior.

We therefore adopt the approach used in IWFVI, which uses the variational distribution as a proposal in a self-normalized importance-sampler. To define an SAA for the objective in this setting, we express the objective at parameters $\phi$ in terms of an expectation with respect to a distribution from the same family with fixed parameters $\tilde{\phi}$,
\begin{align*}
    L_\textsc{f}(\phi) 
    &= 
    \Exp_{p(\cdot|y)}
    \left[
        \log \frac{p(y,z)}{q_\phi(z)} 
    \right]
    =
    \Exp_{q_{\tilde{\phi}}}
    \left[
        w_{\tilde{\phi}}(z)
        \log \frac{p(y,z)}{q_\phi(z)} 
    \right].
\end{align*}
As in standard IWFVI, the importance weights $w_{\tilde{\phi}}(z)$ are intractable, but we can define an SAA for the objective in terms of self-normalized weights, 
\begin{align*}
    \hat{L}_\textsc{f}(\phi \,; \tilde{\phi}) 
    &= 
    \sum_{i=1}^N 
    \hat{w}^{(i)}_{\tilde{\phi}} 
    \log \frac{p(y, z^{(i)})}
              {q_\phi(z^{(i)})}, 
    \qquad
    z^{(i)} \sim q_{\tilde{\phi}}.
\end{align*}
In this surrogate objective, which we will optimize with respect to $\phi$, the quality of the approximation depends on how closely the proposal with parameters $\tilde{\phi}$ matches the posterior. Since our approximation of the posterior will improve during optimization, we will update $\tilde{\phi}$ to the current parameter values $\phi$ at some interval, resulting in an approach that we will refer to as a \emph{sequential} sample-average approximation.

\begin{figure}[t]
    \vspace{-0.5\baselineskip}
    \centering
    \includegraphics[width=\columnwidth]{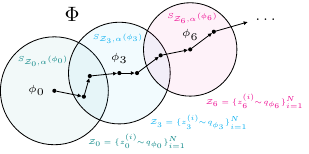}
    \vspace{-1em}
    \caption{Visualization of parameter traces and trust regions corresponding to different SAAs. If after an update $\param \notin S_{\mathcal{Z},\alpha(\saaparam)}$, we set $\saaparam \leftarrow \param$ to construct a new SAA and corresponding trust region.}
    \label{fig:trust_region}
\end{figure}

To determine when we need to generate a fresh SAA, we will define the notion of a trust region. This defines an optimization process in which the SAA is refreshed whenever the optimization trajectory leaves the current trust region. This optimization process is illustrated in Figure~\ref{fig:trust_region} and described schematically in Algorithm~\ref{alg:ours}. We begin by setting the proposal parameters $\tilde{\phi}=\phi_0$ to the initial variational parameters, generating a set of samples $\mathcal{Z} = \{z^{(i)}\}_{i=1}^N$ and defining an SAA of the objective $\hat{L}_\textsc{f}(\phi ; \tilde{\phi})$ and a trust region $S_{\mathcal{Z}, \alpha}(\tilde{\phi})$ based on these samples. We then repeatedly update $\phi_t$ using an optimizer until the value $\phi_t$ no longer lies in the trust region. At this point, we update the proposal parameters $\tilde{\phi} = \phi_t$ and generate a fresh sample set, which we then use to update the SAA and the trust region.

\begin{algorithm}[!t]
    \caption{\ours}
    \label{alg:ours}
    \begin{algorithmic}
        \REQUIRE Initial param. $\param_0$, trust region threshold $\threshold$, data $y$
        \STATE $\saaparam \leftarrow \param_0$ 
        \COMMENT{\hfill \(\triangleright\) Initialize proposal parameter}
        \STATE $\mathcal{Z} \leftarrow \{\sample^{(i)} \sim \vardist_{\saaparam}\}_{i=1}^N$ 
        \COMMENT{\hfill \(\triangleright\) Initialize samples}
        \STATE $\hat{L}_\textsc{f}(\param \,; \saaparam) = \sum_{i=1}^N \hat w^{(i)}_{\saaparam} \log \frac{p(y, z^{(i)})}{q_\phi(z^{(i)})}$
        \COMMENT{\hfill \(\triangleright\) Initialize SAA}        
        \FOR{$t = 1,\ldots,T$}
                \STATE $\param_t = \mathrm{optimizer}\text{-}\mathrm{step}(\hat{L}_\textsc{f}, \param_{t-1})$
            \IF[\hfill \(\triangleright\) Not inside trust region]{$\param_t \notin S_{\mathcal{Z}, \alpha}(\saaparam)$}
                \STATE $\saaparam \leftarrow \param_t$ 
                \COMMENT{\hfill \(\triangleright\) Update proposal}
                \STATE $\mathcal{Z} \leftarrow \{\sample^{(i)} \sim \vardist_{\saaparam} \}_{i=1}^N$ 
                \COMMENT{\hfill \(\triangleright\) Refresh samples}
                \STATE $\hat{L}_\textsc{f}(\param \,; \saaparam) = \sum_{i=1}^N \hat w^{(i)}_{\saaparam} \log \frac{p(y, z^{(i)})}{q_\phi(z^{(i)})}$
                \COMMENT{\hfill \(\triangleright\) Refresh SAA}            
            \ENDIF
        \ENDFOR
    \end{algorithmic}
\end{algorithm}

\vspace{-0.66\baselineskip}
\paragraph{Defining Trust regions.} To define a notion of a trust region, we compute an effective sample size (ESS), which is a proxy measure for the variance of the importance weights. Notably, the importance weights can be decomposed into two parts, (1) the ratio of the variational density and the trust region density, and (2) the ratio between posterior and variational density, 
\begin{align*}
    w_{\saaparam}(\sample) 
    =&  
    v_{\param, \saaparam}(\sample)\:
    w_{\param}(\sample),
    &
    v_{\param, \saaparam}(\sample)
    &=
    \frac{\vardist_{\param}(\sample)}{\vardist_{\saaparam}(\sample)}.
\end{align*}
The variance of $w_\param$ is independent of the fixed proposal parameters and decreases during optimization. 
The variance of $v_{\param, \saaparam}$ measures how similar $\vardist_\saaparam$ is to $\vardist_\param$ and can be controlled by updating the proposal parameters $\saaparam$ to the parameters of the current variational approximation $\phi$ once the ESS drops below a certain threshold. 

We formalize this notion by defining a set-valued function which, for a 
given threshold $\alpha$ and samples $\mathcal{Z} = \{z^{(i)}\}_{i=1}^N$, maps each parameter to a corresponding trust region based on a scoring function, which we choose to be the ESS,
\begin{align*}
    S_{\mathcal{Z}, \alpha}(\saaparam) &= \{\param \in \paramset\ \mid s_\mathcal{Z}(\saaparam, \param) > \alpha \}\\
    s_{\mathcal{Z}}(\saaparam, \param) &= \frac{\left(\sum_{i=1}^N v_{\saaparam, \param}(z^{(i)})\right)^2}{N \sum_{i=1}^N \left(v_{\saaparam, \param}(z^{(i)})\right)^2}.
\end{align*}
In other words, for a given ESS threshold $\alpha$ we can verify $\phi \in S_{\mathcal{Z}, \alpha}(\saaparam)$ by checking $ s_\mathcal{Z}(\saaparam, \param) > \alpha$. We visualize how new trust regions, corresponding to different SAAs, are constructed sequentially during optimization in Figure \ref{fig:trust_region}.

\vspace{-0.5\baselineskip}
\paragraph{Effect of the Trust-Region on Convergence.} For a low enough threshold $\alpha$ the algorithm might converge to an optimal parameter $\hat \phi$ within the trust region of the current SAA that does not yet satisfy our global convergence criteria, i.e. sufficiently minimizes the forward KL-divergence. In these cases, if the variational approximation is not degenerate, we can try recover by decreasing $\alpha$ such that $\hat\phi \notin S_\alpha(\saaparam)$ and continue optimization. As $\alpha \to 1$ the frequency of sample acquisition increases and, for $\alpha = 1$ \ours reduces to standard \baseline. In this work we chose $\alpha$ high enough that the algorithm does not converge prematurely to an optima of an intermediate SAA. In these cases we find that convergence of the training loss can be used as indicator for convergence of the forward KL-divergence or corresponding upper bound, which we verify in our experiments. We also experimented with caching past sample sets to compute a secondary loss based on the last $M$ SAAs. While we did not find it to add additional value in our experiments, it be a useful tool to assess convergence in other settings.

\vspace{-0.5\baselineskip}
\paragraph{Efficient Implementation.}
To avoid recomputing density values for old sample locations,
we cache both sample location $z^{(i)}$ and the corresponding log-joint density of the model $\log p(z^{(i)}, y)$. 
If sampling from the proposal is cheap and memory is of concern, e.g. for large samples set or if past sample sets are stored to compute a validation loss, we can store the random seed instead of the sample and rematerialize the sample when needed.

\section{Related Work}
\paragraph{VI with SAAs}
Recent work that studies SAAs \citep{giordanoBlackBoxVariational2023,burroniSampleAverageApproximation2023} in the context of variational inference focuses on the reparameterized black-box VI setting and optimizes a reverse KL-divergence. 
These methods rely on reparameterization to move samples to areas of high posterior density while keeping a fixed set of noise realizations from the base distribution, which does not depend of the variational parameters. Optimizing a deterministic objective allows the authors to use second-order optimization and linear response methods \cite{giordanoLinearResponseMethods2015} to fit covariances. While these methods allow to realize substantial gains in terms of inference quality and efficiency, in contrast to \ours, they require differentiability of the model. 

\paragraph{Stochastic second-order optimization.}
There is also work outside of the context of SAAs that aims to incorporate second order information to improve stochastic optimization and variational inference. \citet{byrdStochasticQuasiNewtonMethod2016} propose batched-L-BFGS, which computes stable curvature estimates
by sub-sampled Hessian-vector products instead of computing gradient differences at every iteration. This work has also been been adopted to the variational inference setting by \citet{liuQuasiMonteCarloQuasiNewton2021}. 
Pathfinder \cite{zhangPathfinderParallelQuasiNewton2022} uses a quasi-Newton method to find the mode of the a target density and construct normal approximations to the density along the optimization path. The intermediate normal approximations are used to define a variational approximation that minimizes an evidence lower bound. Similar to SAA-based methods, pathfinder can reduce the number of model evaluations by up to an order of magnitude compared to HMC, but requires a differentiable model.

\paragraph{VI with forward KL-divergence.}
\ours is also related to other methods that aim to optimize a forward KL-divergence or its stochastic upper bound. This includes reweighted-wake sleep (and wake-wake) methods \cite{bornscheinReweightedWakeSleep2015, leRevisitingReweightedWakeSleep2018} to which we compare \ours in the experiment section, as well as their doubly-reparameterized variants \cite{tuckerDoublyReparameterizedGradient2018,  finkeImportanceweightedAutoencoders2019, bauerGeneralizedDoublyReparameterized2021}, which are not directly comparable as they require a differentiable model. While the methods above use a single importance sampling step using the variational approximation as a proposal, other methods use more complex proposal including MCMC proposals \cite{naessethMarkovianScoreClimbing2020c,zhangTransportScoreClimbing2023}, approximate Gibbs kernels \cite{wuAmortizedPopulationGibbs2020}, or proposal defined by probabilitic programs \cite{stitesLearningProposalsProbabilistic2021, zimmermannNestedVariationalInference2021a}. While these methods do not necessarily require a differentiable proposal they are not designed to be sample efficient but to approximate complex target densities.

~
\begin{figure*}[t!]
    \centering
    \includegraphics[width=1.\textwidth]{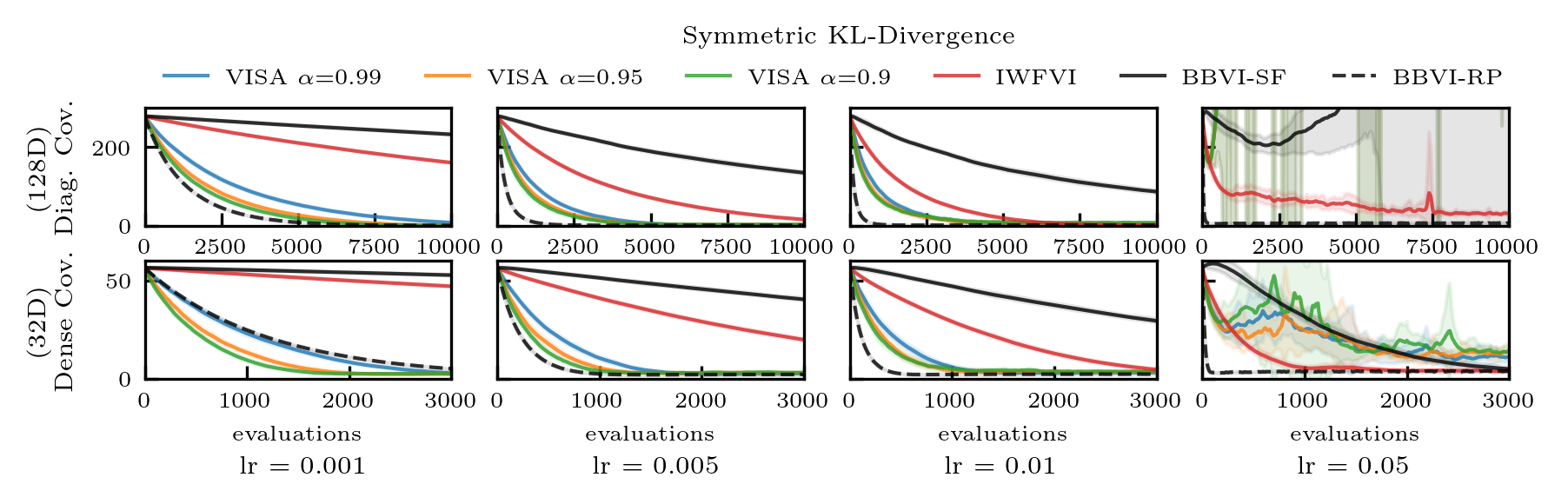}
    \vspace{-0.5cm}
    \caption{Symmetric KL-divergence as a function of the number of model evaluations for a Gaussian target with diagonal covariance matrix (top row) and dense covariance matrix (bottom row). For small learning rates (0.001, 0.005, 0.01) \baseline and BBVI-SF, need a larger number of model evaluations to converge. \ours converges much faster as it compensates for the small step size by reusing samples. For a learning rate of 0.05 \ours fails to reliably converge, while \baseline still converges. 
    Overall, \ours converges faster or at the same rate as \baseline and BBVI-SF with the same or higher learning rates.
}
    \label{fig:normals_convergence}
\end{figure*}

\section{Experiments}
We compare \ours to standard importance-weighted VI (\baseline) in terms of inference quality and the number of model evaluations. We assess inference quality by measuring or estimating the forward KL-divergence to the true posterior, or when such an estimate is not available by assessing the log-joint probability of samples under the model. In the Gaussian experiment we also compare to black-box VI, which optimizes the reverse KL-divergence, and therefore report inference quality by measuring the symmetric KL-divergence to allow for a fair comparison.

\subsection{Gaussians}
To study the effect of different learning rates and ESS threshold parameters, we first evaluate \ours on approximating medium- to high- dimensional Gaussians and compare the inference performance over the number of model evaluations to \baseline, standard reparameterized variational inference (BBVI-RP) and variational inference using a score-function gradient estimator (BBVI-SF).  
Notably, we include BBVI-RP as a reference only, showcasing that faster convergence can be achieved by leveraging the differentiability of the model, and do not compare to it directly.
To allow for a fair comparison between methods that optimize a forward KL-divergence (\ours, baseline) and methods that optimize a reverse KL-divergence (BBVI), we evaluate the inference qualify in terms of the symmetric KL-divergence.

We study two different target densities, a (1) $D=128$ dimensional Gaussian with a diagonal covariance matrix and (2) $D=32$ dimensional Gaussian with a dense covariance matrix. 
For the diagonal covariance matrix, we scale the variances of the individual dimension from $\sigma_{\min}=0.1$ to $\sigma_{\max}=1$ such that the covariance matrix takes the form
\begin{align*}
    C_\mathrm{diag} = 
    \mathrm{diag}
    \left(\left[
        \sigma_{\min} + (i-1)*\frac{\sigma_{max} - \sigma_{min}}{D-1}
        \right]_{i=1}^D\right)
.
\end{align*}
To create the dense covariance matrix we first sample a random positive semi-definite matrix $M = AA^T$, where $A_{ij} \sim \mathcal{U}(0, 1)$ and then construct the covariance matrix
\begin{align*}
    C_\mathrm{dense} = \left(\frac{M}{\vert| M \vert|_\mathrm{F}} + 0.1 \mathbb{I}\right)
    .
\end{align*}

Figure \ref{fig:normals_convergence} shows the results for different learning rates $\text{lr} \in \{0.001, 0.005, 0.01, 0.05\}$ and ESS threshold parameters $\alpha \in \{0.9, 0.95, 0.99\}$. We compute gradient estimates for \ours, \baseline, and BBVI-SF with $N=10$ samples, and gradient estimates for BBVI-RP using a single sample. 

We observe that \ours converges substantially faster than IWFVI and BBVI-SF at lower learning rates (0.001, 0.005, 0.01) for both, targets with diagonal- and dense covariance matrix. The difference in the convergence rate gets less pronounced as the learning rate increases. For large learning rates (0.05) \ours fail to converge reliably, while \baseline still converges. For even higher learning rates all methods but BBVI-RP fail to converge. 
If we compare convergence across learning rates, we observe that \ours converges faster or at the same rate as \baseline with the same or higher learning rate.
%
In the Gaussian experiment, the overall effect of the threshold $\alpha$ on the convergence rate is minor, but we observe slightly faster convergence for lower ESS thresholds. 
We also find that, especially for lower ESS thresholds, \ours is also more susceptible to underestimating posterior variance and might not fully converge in the final phase of training. We hypothesize that this is due to the deterministic optimization procedure, which overfits to high probability samples if the samples are not refreshed frequently enough. A possible remedy for this is to increase the ESS threshold during the final phase of training, for $\alpha \to 1$ this recovers the behaviour of \baseline. 
However, since the overall effect of the threshold on the rate of convergence is small, a viable strategy is to employ a fixed threshold value that is somewhat conservative, such as $\alpha=0.95$ or $\alpha=0.99$. 

Overall we find that \ours converges faster or at the same rate as \baseline and BBVI-SF for a range of different learning rates and ESS thresholds. Moreover \ours is substantially more robust to the choice of learning rate. Using a deterministic objective allows us to draw fresh samples less frequently and consequently requires less evaluations of a potentially expensive to evaluate model log-joint density.

\subsection{Lotka-Volterra}
The Lotka-Vorterra predator-prey population dynamics \cite{lotkaPrinciplesPhysicalBiology1925, volterraFluctuationsAbundanceSpecies1927} are modeled by a pair of first-order ordinary differential equations (ODEs),
\begin{align*}
    \frac{du}{dt} = (\alpha - \beta v)u,
    && 
    \frac{dv}{dt} = (-\gamma + \delta u)v,
\end{align*}
where $v$ denotes the predator- and $u$ denotes the prey population. We will in the following denote the pair of predator-prey populations at time $t$ with $z_t=(u_t, v_t)$.
The dynamics of the ODE are governed by its respective population growth and shrinkage parameters $\theta = (\alpha, \beta, \gamma, \delta)$, which we would like to infer together with the initial conditions of the system given noisy observations $y_{1:T} = (y_1, \ldots, y_T)$.

\begin{figure*}[t!]
    \includegraphics[width=\textwidth]{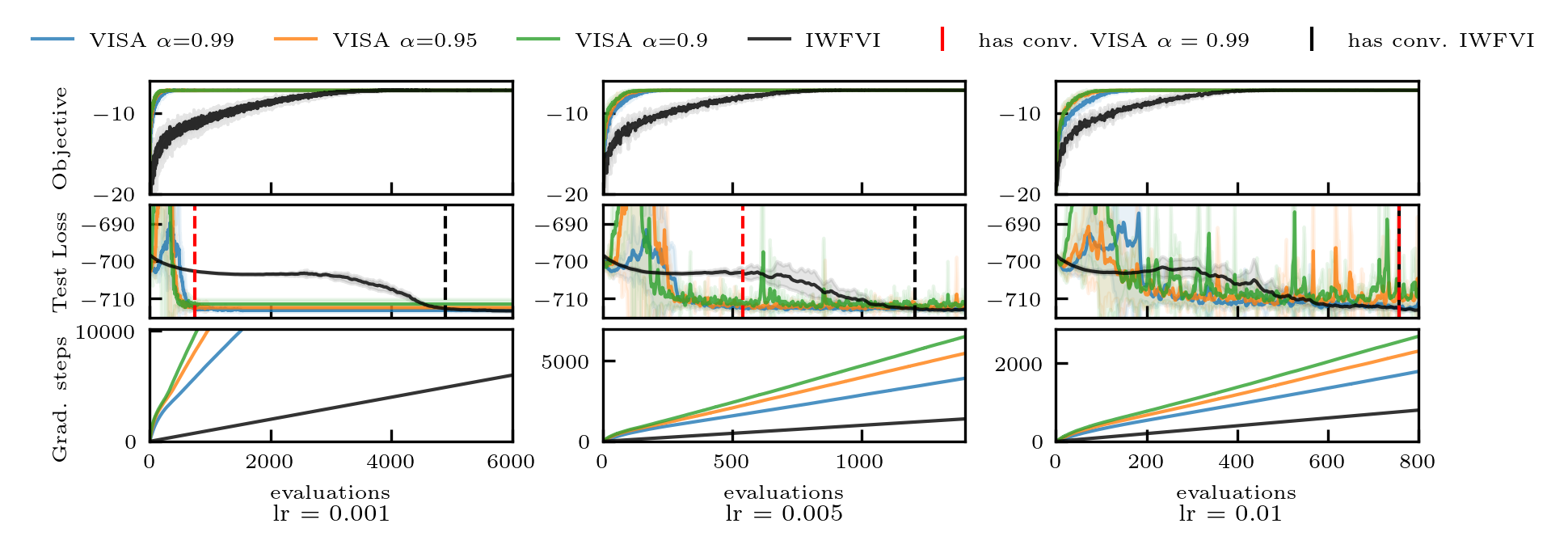}
    \vspace{-0.7cm}
    \caption{Results for Lotka-Volterra model with different learning rates. (Top row) Training objective over number of model evaluations. (Middle row) Approximate forward KL-divergence computed on reference samples obtained by MCMC. For smaller step sizes (0.001, 0.005) \ours achieves comparable forward KL-divergence to \baseline while requiring significantly less model evaluations to converge (see vertical lines). For larger step sizes (0.01) \ours only converges with a high ess threshold (0.99) for which it requires approximately the same number of evaluations as \baseline. (Bottom row) Gradient steps over number of batch evaluations of the model, each batch evaluation corresponds to evaluating a batch of $N=100$ samples. \ours requires fewer evaluations per gradient step compared to \baseline.
    }
    \label{fig:lotka-volterra-kl-over-evals}
\end{figure*}

Following \citet{carpenterPredatorPreyPopulationDynamics2018}, we place priors over the initial population size $z_0$ and system parameters $\theta$
\begin{align}
    \label{eq_lotka_volterra_prior}
    z^\mathrm{prey}_0, z^\mathrm{pred}_0 &\sim \mathrm{LogNormal}(\log(10), 1), \\
    \alpha, \gamma &\sim \mathrm{Normal}(1, 0.5), \\
    \beta, \delta &\sim  \mathrm{Normal}(0.05, 0.05)
    ,
\end{align}
and assume a fractional observation error, 
\begin{align*}
    y^\mathrm{prey}_t, y^\mathrm{pred}_t &\sim \mathrm{LogNormal}(\log z_t, \sigma_t) \\
    \sigma_t &\sim  \mathrm{LogNormal}(-1, 1)
    .
\end{align*}
Given an initial population $z_0$, system parameters $\theta$, and observations $y_{1:T}$ we can solve the ODE numerically to obtain approximate population sizes $z_{1:T}$ for time steps $1, \ldots, T$ which we use to compute the likelihood of the observed predator-prey populations,
\begin{align*}
    p(y_{1:T} \mid z_0, \theta) &= \prod_{t=0}^{T} p(y_t \mid z_t)
    .
\end{align*}
Our goal is to learn an approximation to the posterior $p(\theta, z_0 \mid y)$ by minimizing the evidence upper bound
\begin{align}
    \label{eq:upper-bound-lotka-volterra}
    L_\textsc{f}(\phi) :=
    \mathbb{E}_{(z_0, \theta) \sim p(\cdot, \cdot \mid y)}\left[
    \log
    \frac{
        p(z_0, y, \theta)
    }{
        q_\phi(z_0, \theta)
    }
    \right]
    .
\end{align}
We model the variational approximation $q_\phi$ for the interaction parameters $\theta$ and initial population sizes $z_0$ as jointly log-normal and initialize $\phi$ such that the the marginal over $z_0$ matches the prior (Equation~\ref{eq_lotka_volterra_prior}) and the marginal over $\theta$ has similar coverage to the prior. We approximate the objective and its gradient using with $N=100$ samples.
To specify a common convergence criteria, we compute the highest common test loss value for \ours with $\alpha = 0.99$ and \baseline that is not exceeded by more than $1$ nat by all consecutive test loss values. The convergence threshold computed this way is $-712.6$ nats.

To evaluate the inference performance, we first generate $N=10\,000$ approximate posterior samples using a No-U-Turn Sampler (NUTS) \cite{hoffmanNoUTurnSamplerAdaptively2014a} with $10\,000$ burn-in steps and window-adaption, which generally provides good performance out of the box\footnote{NUTS is an adaptive Hamiltonian Monte Carlo sampler and uses the gradient information of the model to guide the generation of proposals. As such it requires the log-joint density model to be differentiable which is not required by \ours or \baseline.}. 
We use the approximate posterior samples to approximate an ``oracle'' for the upper bound in Equation \ref{eq:upper-bound-lotka-volterra},
\begin{align*}
    \hat{L}^\textsc{nuts}_\textsc{f}(\phi) 
    =
    \frac{1}{N}
    \sum_{i=1}^N
    \log
    \frac{
        p(z^{(i)}_0, y, \theta^{(i)})
    }{
        q_\phi(z_0^{(i)}, \theta^{(i)})
    },
\end{align*}
which we evaluate along with the training objective during optimization to assess convergence.

We find that \ours is able to obtain variational distribution of similar quality to \baseline while requiring significantly fewer model evaluations for smaller learning rates (see Figure~\ref{fig:lotka-volterra-kl-over-evals}).
Interestingly, \ours requires significantly fewer model evaluations per gradient step during the early stages of training, while requiring slightly more evaluations per gradient step thereafter. We hypothesise that this is again a result of under approximating posterior variance in the later stages of training. As a result, even small changes in the variational distribution can lead to big changes in the ESS, which triggers the drawing of fresh samples.
For \ours, we also find a more pronounced difference between different ESS thresholds and their influence on convergence. Runs with a higher ESS thresholds converge more stably and are able to achieve lower test loss in the final stages of training.

\subsection{Pickover Attractor}

\begin{figure*}
    \centering
    \includegraphics[width=\textwidth]{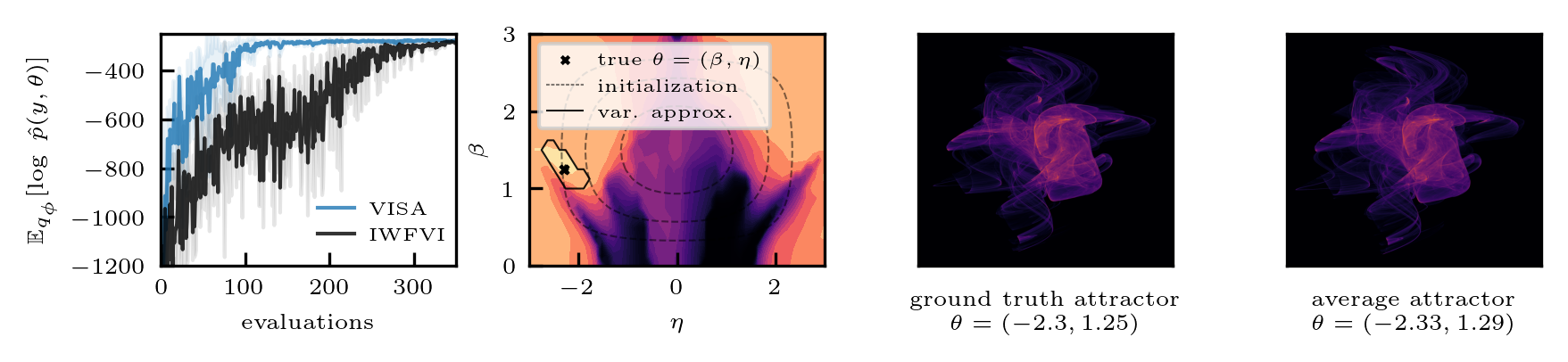}
    \vspace{-0.5cm}
    \caption{Results for Pickover attractor. (a) Approximate log-joint density over number of batch-evaluations of model. (b) Log-joint approximation plotted over domain of prior. The variational approximation capture the high density area containing the data. (c) Visualization of pickover attractor with ground truth parameters $\theta=[-2.3, 1.25]$. (d) Visualization of attractor with average system parameters computed over $10.000$ samples from the learned variational approximation.
    Each evaluation in the plot corresponds to evaluating a batch of $N=10$ samples.
    }
    \label{fig:experiment_pickover}
\end{figure*}
\vspace{\baselineskip}

Following \citet{rainforthBayesianOptimizationProbabilistic2017}, we model a 3D Pickover attractor \cite{pickoverPatternBookFractals1995} with parameters $\theta = (\beta, \eta)$,
\begin{align*}
    x_{t+1, 1} &= \sin(\beta x_{t, 2}) - \cos(2.5 x_{t,1}) x_{t,3}\\
    x_{t+1, 2} &= \sin(1.5 x_{t, 1})x_{t,3} - \cos(\eta x_{t,2})\\
    x_{t+1, 3} &= \sin(x_{t, 1})
    .
\end{align*}
Due to its chaotic nature the system is sensitive to small perturbations in its initial state, i.e. even small variations in the initial state lead to exponentially fast diverging trajectories. Therefore, to track the evolution of the system, we employ a bootstrap particle filter \cite{gordonNovelApproachNonlinear1993} which assumes noisy observations $y := y_{1:T}$ and introduces auxiliary variables $z := z_{1:T}$ to model the latent state of the system. 
We define the prior over system parameters 
\begin{align*}
    p(\theta) = 
    \begin{cases}
        1/18 & -3 \leq \theta_1 \leq 3, 0 \leq \eta \leq 3\\
        0 & \text{otherwise}
    \end{cases}
    ,
\end{align*}
and model the transition and observation as, 
\begin{align*}
    z_0 &\sim \mathcal{N}(\cdot \mid \mathbf{0}_3, \mathbb{I}_3),&\\
    z_t &\sim \mathcal{N}(\cdot \mid h(z_{t-1}, \theta), \sigma_z)\quad &\text{for}\ t>0,\\
    y_t &\sim \mathcal{N}(\cdot \mid z_{t}, \sigma_y)\ &\text{for}\ t\geq0,
\end{align*}
where $\sigma_z=0.01$, $\sigma_y=0.2$, and $h$ evolves the system by one time step using the equations of the Pickover attractor described above. The particle filter is used to simulate $T=100$ time steps with $M=500$ particles, which renders evaluating the model expensive.

To restrict the proposal to the same domain as the prior by first sampling from a Normal, which we denote $\bar q_\phi$ and parameterized with a mean and lower Cholesky factor, and then transform the sample by $$f(\theta) = (\tanh(3\cdot\theta_1), \tanh(1.5 \cdot \theta_2 + 1.5)).$$ The density of the transformed samples is $q_\phi(\theta) = \bar{q}_\phi(\theta)\lvert\mathrm{det}\frac{d f(\theta)}{d\theta}\rvert^{-1}$, which is appropriately restricted to the domain of the prior.

We are interested in approximating the marginal posterior $p(\theta \mid y)$ over system parameters by optimizing the evidence upper bound
\begin{align*}
    \Exp_{p_\theta(\theta \mid y)}\left[
        \log \frac{p_\theta(y, \theta)}{q_\phi(\theta)}
    \right]
    &=
    \Exp_{p_\theta(\theta \mid y)}\left[
        \log 
        \frac{
            \mathbb{E}_{p_{_\mathrm{pf}}}[\hat p(y \mid \theta)]
        }{ 
            q_\phi(\theta)
        }
    \right]
    \\
    &\leq
    \Exp_{p_\theta(\theta \mid y)}\left[
        \mathbb{E}_{p_{_\mathrm{pf}}}\left[
            \log 
            \frac{
               \hat p(y \mid \theta)
            }{ 
                q_\phi(\theta)
            }
        \right]
    \right]
    .
\end{align*}
To obtain a tractable objective we replace the intractable marginal likelihood $p(y\mid \theta) \approx \hat p_\theta(y \mid \theta)$ with the marginal likelihood estimate obtained by running the particle filter \cite{naessethElementsSequentialMonte}, similar to pseudo-marginal methods \cite{andrieuParticleMarkovChain2010} and approximate the gradient with $N=10$ samples. As the likelihood estimate is non-differentiable due to the discrete ancestor choices made inside the particle filter, we cannot run NUTS to obtain approximate posterior sample as before, but instead report the log-joint density of the variational distribution. 

We observe that \ours converges more stably with fewer samples compared to \baseline and find that attractors corresponding to samples from the variational approximation look qualitatively similar to those based on the true parameters.
We summarize the result in Figure. \ref{fig:experiment_pickover}.

\section{Discussion and Limitations}

In this paper we developed \ours, a method for approximate inference for expensive to evaluate models that optimizes the forward KL-divergence through a sequence of SAAs.
Each SAA is optimized deterministically and requires and fixes a single set of samples, hereby requiring new model evaluations only when the SAA is refreshed. To track the approximation quality of the current SAA, \ours computes the ESS of the ratio between the current variational distribution and the proposal distribution that was used to construct the SAA. If the ESS falls below a predefined threshold, a new SAA approximation is constructed based on fresh samples from the current variational distribution. We observe gains of a factor 2 or more in terms of the number of required model evaluations for conservatively chosen step sizes, while achieving similar posterior approximation accuracy as \baseline, the equivalent method that does not employ the sequential sample-average approximation.

\paragraph{Underapproximation of posterior variance.}
Both reparameterized VI, which optimizes the reverse KL-divergence, and \baseline, which optimizes the forward KL-divergence via importance sampling, are prone to under approximating posterior variance. In the case of reparameterized VI, this can often be attributed to the mode seeking behaviour of the reverse KL-divergence, while in \baseline the low effective samples sizes can lead to over fitting to a small number of high-weight samples. We found that keeping the samples fixed for too long, i.e.~using an ESS threshold that is too low, can exacerbate this problem, as the optimizer can take multiple steps towards the same high-weight samples. 

\citet{giordanoBlackBoxVariational2023} and \citet{burroniSampleAverageApproximation2023} showed that when applying SAA to reparameterized VI, it is possible to make use of second-order methods. We experimented with optimizing SAAs with L-BGFS, which is a quasi-Newton method with line search. However, we found that in the setting of optimizing a forward KL with relatively few samples, L-BGFS can amplify the problem of overfitting, often leading to instabilities and collapsed variational distributions. 

\paragraph{Number of latent variables and parameters.}
Because \ours employs a relatively small number of samples and does not refresh samples at every iteration, we found that it is not well-suited to models with a large number of latent variables or large number of parameters. This agrees with theoretical findings by \citet{giordanoBlackBoxVariational2023}, who show that SAAs for a full covariance Gaussian fail if the number of samples is not at least in the same regime as the number of latent dimensions. 
\citet{burroniSampleAverageApproximation2023} manage to train full covariance normal approximation by using a sequence of SAAs using increasingly large sample sizes, however, this is directly opposed to our goal of reducing the number of model evaluations in expensive to evaluate models.


\section*{Acknowledgement} The authors would like to thank Tamara Broderick for helpful discussion about sample-average approximations in the context of reparameterized variational inference.

\bibliography{main}
\bibliographystyle{icml2024}
\end{document}